\documentclass[letterpaper, 10 pt, conference]{ieeeconf}  % Comment this line out if you need a4paper

\IEEEoverridecommandlockouts                              % This command is only needed if 
\usepackage{graphicx}
\usepackage{bm}
\usepackage{amsmath} %
\usepackage{amssymb}  %
\usepackage{multirow}
\usepackage{todonotes}
\setuptodonotes{inline}
\usepackage{url}
\usepackage{algorithm}
\usepackage{algpseudocode}
\usepackage{siunitx}
\usepackage[hidelinks]{hyperref}
\usepackage{cleveref}
\usepackage{booktabs}

\usepackage[style=ieee, doi=false, isbn=false, url=false]{biblatex} %Imports biblatex package
\AtBeginBibliography{\small}
\title{\LARGE \bf
Replicating Human Anatomy with Vision Controlled Jetting -- \\
A Pneumatic Musculoskeletal Hand and Forearm
}

\author{Thomas Buchner$^{1\dagger}$, Stefan Weirich$^{1\dagger}$, Alexander M. Kübler$^{1}$, \\Wojciech Matusik$^{2,3}$, and Robert K. Katzschmann$^{1*}$ 
\thanks{$^{1}$Soft Robotics Lab, IRIS, D-MAVT, ETH Zurich, Switzerland}%
\thanks{$^{2}$Inkbit, 1 Cabot Rd, Suite 400, Medford 02155 MA, USA}%
\thanks{$^{3}$CSAIL, MIT, 32 Vassar St, Cambridge, 02139 MA, USA}%
\thanks{$\dagger$ Thomas Buchner and Stefan Weirich are co-first authors.}
\thanks{$*$ Corresponding author: \href{mailto:rkk@ethz.ch}{\tt rkk@ethz.ch}}
}

\begin{document}
\maketitle
\thispagestyle{empty}
\pagestyle{empty}

\begin{abstract}
The functional replication and actuation of complex structures inspired by nature is a longstanding goal for humanity.
Creating such complex structures combining soft and rigid features and actuating them with artificial muscles would further our understanding of natural kinematic structures.
We printed a biomimetic hand in a single print process comprised of a rigid skeleton, soft joint capsules, tendons, and printed touch sensors. We showed it's actuation using electric motors.
In this work, we expand on this work by adding a forearm that is also closely modeled after the human anatomy and replacing the hand's motors with 22 independently controlled pneumatic artificial muscles (PAMs).
Our thin, high-strain (up to 30.1\,\%) PAMs match the performance of state-of-the-art artificial muscles at a lower cost.
The system showcases human-like dexterity with independent finger movements, demonstrating successful grasping of various objects, ranging from a small, lightweight coin to a large can of 272\,g in weight.
The performance evaluation, based on fingertip and grasping forces along with finger joint range of motion, highlights the system’s potential.
\end{abstract}

\section{Introduction}

\subsection{Motivation}
When we engage in routine activities like sipping from a bottle, composing a letter, or typing out a document, the intricate movements and manipulations required come naturally to us. This seamless integration of thought and action stems from our cognitive faculties and the exceptional dexterity of human hands~\cite{faisal_manipulative_2010}. The human hand distinguishes itself through its combination agility, strength, and speed. Current robotic hands fall behind in those metrics.
The design of robotic and prosthetic hands has been researched for many decades but still has not reached the levels of dexterity, versatility, and compactness of their human counterpart~\cite{kadalagere_sampath_review_2023}. 

\begin{figure}[t]
    \includegraphics[width=\columnwidth]{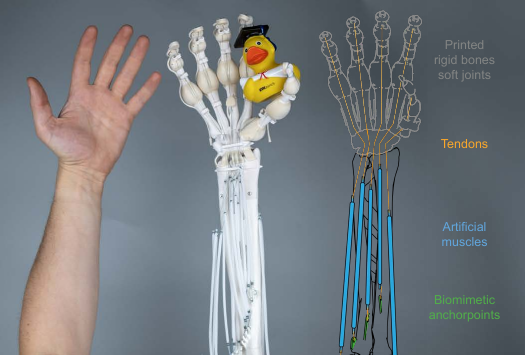}
    \centering
    \caption{\textbf{Pneumatic musculoskeletal hand and forearm.} Inspired by the human anatomy, we printed a musculoskeletal hand and forearm. The soft-rigid hybrid hand is printed in a single process and does not need assembly apart from attaching thin McKibben muscles for actuation. We replicated the anchor points and muscle lengths from the biological inspiration.}
    \label{Fig:EyeCatcher}
\end{figure}

Previously developed robotic hands are either biomimetic and very complex to fabricate, or the designs use strong simplifications for ease of fabrication and cost reduction. 
Biomimicry and dexterity versus the ease of fabrication and cost efficiency is a critical trade-off decision in robotic and prosthetic hand design to date.

\begin{figure*}[t]
    \includegraphics[width=\textwidth]{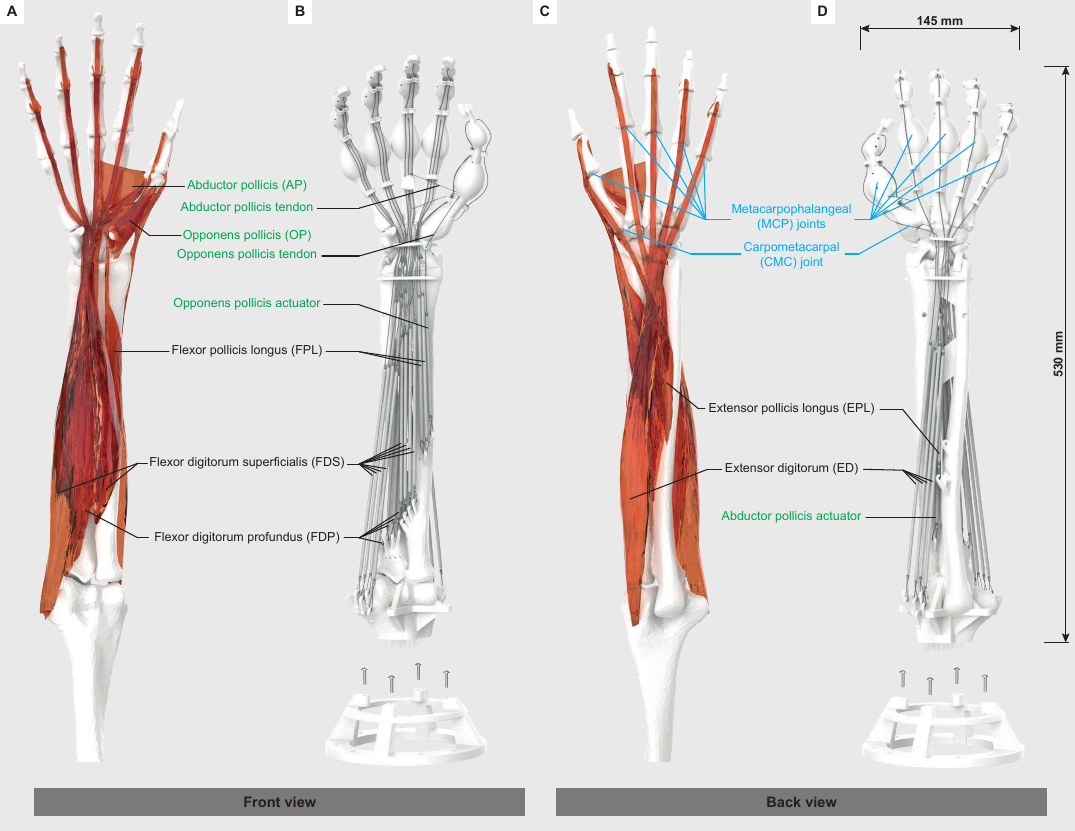}
    \centering
    \caption{\textbf{Inspiration from human hand and forearm replicated in a 3D print.} \textbf{A}~Front view of a 3D bone model of a human arm and hand with accurate muscle placement. \textbf{B}~Front view of a 3D model of the printed hand and forearm. The bone shapes and positions were extracted from MRI data. The tendons and actuators replacing the muscles are placed according to their biological counterparts' anchor points and positions. \textbf{C}~Back view of model shown in sub-figure A. \textbf{D}~Back view of model shown in sub-figure B. The sperical volumes shown around the joints are our modeled joint capsules/ligaments.}
    \label{Fig:BiologicalInspiration}
\end{figure*}

\subsection{Related Work}
Biomimetic hand designs~\cite{zhe_xu_design_2016,tasi_design_2019} mimic human bones, joints, tendons, and ligaments using tendon-driven mechanisms. While the hands themselves achieve high levels of biomimicry, the biomimetic aspects in the forearm area are neglected entirely by using rotating electric motors to actuate the tendons. Fabrication and assembly of these hands are complex and time-consuming. So far, no biomimetic robotic hand has been realized in combination with an equally biomimetic forearm and wrist.

Many robotic hands use non-biomimetic, standard mechanical components, such as cables, gears, belts, linkages, and parallel mechanisms, or directly integrate electric motors into the fingers~\cite{piazza_century_2019,nazma_tendon_2012,lee_development_2009,kashef_robotic_2020}.
The electric motors do not resemble the features of human muscles and constrain the compactness and compliance of the hand design.  The motor's rigid components can become unsafe in dynamic, collaborative applications~\cite{firth_anthropomorphic_2022}.
Recent hand designs place the electric motors either in the forearm~\cite{zhe_xu_design_2016,tasi_design_2019,grebenstein_hand_2012} or in the palm~\cite{liow_olympic_2020}. The motor arrangements in the forearm enable accurate control of complex hands with a large number of DoF but come with major drawbacks such as the weight, size, and bulkiness of the design, often limiting possible applications. The number of pneumatic artificial muscles (PAMs) that can be located in the palm is limited due to spatial constraints.
Commercially available robotic~\cite{ruehl_experimental_2014} and prosthetic~\cite{haverkate_assessment_2016,azeem_study_2022} hands typically use electric motors for actuation and implement varying design simplifications with conventional mechanical components.
Manipulation research also focuses on unteractuated designs resulting in reduced dexterity. Catalano \textit{et al.}~\cite{catalano_adaptive_2014} designed a five-fingered robotic hand with only one motor using adaptive synergies. 

In soft robotics, robotic hands typically integrate bending PAMs in the fingers~\cite{fras_soft_2018,truby_soft_2019,zhou_bcl-13_2018}. Functional soft prosthetic hands with bending PAMs~\cite{gu_soft_2021} and soft, tendon-driven hands made from foam~\cite{king_design_2018} or 3D printed thermoplastic polyurethane~\cite{mohammadi_practical_2020} have been realized. While this actuation method has advantages for many applications requiring safety and compliance, it inherently limits the dexterity, payload capacity, and level of biomimicry in the hand design.

The biological design -- consisting of rigid bones, soft muscles, tendons, ligaments, and compliant skin -- inspired the development of hybrid soft-rigid robots (\Cref{Fig:EyeCatcher}). While the rigidity of such a robot provides a stiff underlying structure for enhanced strength, the softness enables inherent system compliance that is sought after in many areas of robotics. Such robots have the great potential to be placed within human environments and provide smoother interactions without having to fear harm caused by the robot. %Especially for applications such as grasping and heavy, dexterous object manipulation, a hybrid soft-rigid robot could solve canonical challenges.

A great option to drive tendons in hybrid soft-rigid robots are soft linear artificial muscles that are compliant and can stretch and contract in one direction.
Soft linear artificial muscles can stretch and contract in one direction, making them suitable for driving tendons in hybrid soft-rigid robots. Air pressure is a common method for actuating these muscles. The McKibben, a well-known type of pneumatic muscle, increases its diameter when pressurized, leading to contraction due to its woven outer layer~\cite{chou_measurement_1996}. Other pneumatic designs include high-strain inverse artificial muscles~\cite{hawkes_design_2016} or pouch motors~\cite{niiyama_pouch_2015}.
Other artificial muscles include using shape memory alloys~\cite{park_novel_2020}, dielectric elastomer actuators (DEAs)~\cite{youn_dielectric_2020}, and hydraulically amplified self-healing electrostatic (HASEL) muscles~\cite{rothemund_hasel_2021}. Compared to pneumatic muscles, these alternatives are directly actuated with electric energy, requiring less equipment to operate. However, pneumatically driven muscles can respond faster than shape memory alloys and generate higher forces than DEAs and achieve larger contractions than HASEL actuators~\cite{singh_controlled_2022}. 

Recently, the integrated fabrication of soft and rigid structures was simplified by advances in multi-material additive manufacturing. Vision controlled jetting (VCJ), a contactless 3D inkjet deposition method, can print slow-curing polymers like thiol-ene. These polymers are less viscous and more stable to environmental influences than state-of-the-art acrylate materials~\cite{buchner_vision-controlled_2023}. Using wax-based support material that easily melts at low temperatures allows for thin-walled cavities and long hollow tubes following a complex path to be printed in place. These features enable the printing of complete and functional systems with minimal post-processing and assembly. 

\subsection{Contributions}
The contributions of this work are: 
\begin{itemize}
    \item The design of an accurate biomimetic hand and forearm actuated with linear PAMs.
    \item An easy-to-manufacture, cost-effective, high-strain PAM.
    \item The integration of vision controlled jetting parts into a functioning robotic system.
    \item The evaluation of the system.
\end{itemize}

\section{Methods}

\subsection{Anatomy of the Human Hand and Forearm}
In the human hand, the movements of the fingers are actuated by two separate muscle groups~\cite{schunke_prometheus_2005, zilles_anatomie_2010}: the extrinsic muscles located in the forearm and the intrinsic muscles located in the palm. Extrinsic muscles are responsible for larger movements of the fingers. They can be further categorized into two functional groups: flexors, which bend the fingers, and extensors, which straighten the fingers.
\Cref{Fig:BiologicalInspiration} shows the most important human muscles and joints with their acronyms.
These intrinsic muscles spread the fingers and contribute to the flexion of the MCP joint, the base joint between the fingers and the palm.
The thumb is actuated independently from the other fingers by four main extrinsic muscles, which are important for gripping and pinching motions. For the opposing thumb motion used for many grasps, an important role is played by the CMC joint of the thumb and intrinsic muscles OP and AP.
Partially shared muscle bellies for different fingers and interconnected tendons add additional complexity. This is why full flexion or extension of an individual finger is usually difficult without the muscle affecting the neighboring fingers~\cite{tasi_design_2019}.

Human skeletal muscles exhibits a maximum static stress of \qty{0.35}{\mega\pascal}, which can only be held briefly due to muscle fatigue. The maximum sustainable stress is around \qty{0.1}{\mega\pascal}~\cite{hunter_comparison_1992} and the skeletal muscle strain is typically around \qty{20}{\percent}~\cite{mirvakili_artificial_2018}. Muscles can only generate tensile forces and are typically found in antagonist arrangements to enable flexion and extension motions.

\begin{table}[t]
    \centering
    \caption{Muscles imitated by PAMs.}
    \label{tab:Muscles}
    \begin{tabular}{llr}\toprule
    \textbf{\begin{tabular}{@{}c@{}} Acronym\end{tabular}} & 
    \textbf{\begin{tabular}{@{}l@{}} Reference \\ muscle\end{tabular}} & 
    \textbf{\begin{tabular}{@{}r@{}} Actuator \\ amount x length\end{tabular}}\\
    \midrule
    Flexor Digitorium  Profundus    & FDP   & $4 \times \qty{220}{\milli\meter}$\\
    Flexor Digitorium Superficialis & FDS   & $4 \times \qty{200}{\milli\meter}$\\
                                    &       & $4 \times \qty{160}{\milli\meter}$\\
    Flexor Pollicis Longus          & FPL   & $2 \times \qty{160}{\milli\meter}$\\
    Extensor Digitorum              & ED    & $4 \times \qty{220}{\milli\meter}$\\
    Extensor Pollicis Longus        & EPL  & $1 \times \qty{160}{\milli\meter}$\\
    Abductor Pollicis               & AP    & $2 \times \qty{160}{\milli\meter}$\\
    Opponens Pollicis               & OP    & $1 \times \qty{140}{\milli\meter}$\\ \bottomrule
    \end{tabular}
\end{table}

\subsection{Forearm Design}
We designed the hand based on a detailed 3D anatomical model of a human male right forearm, derived from MRI scans from the BodyParts3D database ~\cite{mitsuhashi_bodyparts3d_2009}. And characterized the hand with electric motors~\cite{buchner_vision-controlled_2023}. The anatomical bones model established the rigid basis of our design, while muscle and tendon models were used as inspiration for placing anchor points for attaching artificial muscles and shaping the soft joints and material transitions.
\Cref{Fig:BiologicalInspiration} shows the integrated PAMs and joints, and \Cref{tab:Muscles} lists the PAMs.

The main focus of this work was on recreating the functionality of the extrinsic muscles ED, FDS, and FDP, which effectuate the larger finger movements for flexion and extension motions. The resulting system has a total of 22 PAMs. We neglected inter-dependencies between muscles, as found in nature, to reduce mechanical and control complexity. It could, however, be realized by appropriate control strategies.

Special consideration was given to the wrist and thumb CMC joints due to their complexity and the necessity of seamless integration of soft and rigid materials. In the wrist, we replicated seven individual tendons, navigating the challenge of connecting five different bones. 

For the thumb CMC joint, a saddle joint, we modified adjacent bones to create a smooth surface for the joint capsule, also modeling soft joints and reinforcing ligaments for the remaining two interphalangeal joints of the thumb.
To enable an opposing thumb motion, we mimic the FPL and EPL muscles, as well as the intrinsic AP and OP muscles. The hand does not provide enough space to place PAMs of significant length at the original location of the intrinsic muscles in the palm. Therefore, the corresponding PAMs are placed in the forearm area. The tendons are guided through tunnels in the metacarpal bone and along the carpal base to recreate the direction of action of these muscles.

To facilitate the fixation of the fishing line tendons at the fingers, we designed attachment holes within the finger bones. To guide the air supply tubes of the PAMs, we integrate tunnels through the bones.
The forearm was printed as a single rigid component, with the elbow joint stiffened and bones fused at the elbow. %Additional stability was achieved by connecting the radius and ulna with thin rigid links, inspired by the interosseous membrane.

We use vision controlled jetting, a contactless 3D inkjet deposition method, to fabricate our soft-rigid-hybrid hand. We were able to print the hand in one process with sensor pads and pressure lines ready to use. We added tendons made from fishing line to conduct experiments.

\subsection{Pneumatic Artifical Muscle}

\begin{figure}[t]
    \includegraphics[width=\columnwidth]{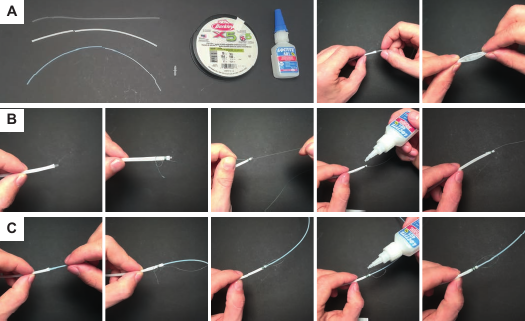}
    \centering
    \caption{\textbf{Materials and step-by-step illustration of the manual fabrication process for the PAM.} \textbf{A} Overview of the used materials and the insertion of the tube into the braided sleeving. \textbf{B} Manufacturing steps for the end terminal. \textbf{C} Manufacturing steps for the end part with the air supply inlet.}
    \label{Fig:PAM-fab}
\end{figure}

We fabricated thin PAMs based on the McKibben principle~\cite{chou_measurement_1996}. \Cref{Fig:PAM-fab} shows the manufacturing process. The PAMs consist of Flexo PET 3.2 (Techflex Inc.) braided sleeve made of polyethylene terephthalate fibers and an expandable silicone tubing (Conta Elastomers GmbH) with an inner diameter of \qty{1.5}{\milli\meter}, wall thickness of \qty{0.25}{\milli\meter}, elongation at break of \qty{540}{\percent}, and hardness Shore A30. 
The silicone tube is inserted into the braided sleeving. We use 5-braid (polyethylene) fiber fishing line (Pure Fishing Inc.) to close the ends with a knot. We close the knot with a Loctite 401 adhesive (Henkel \& Cie. AGA) to ensure airtight sealing and avoid knot slipping. The fishing line on the upper end of the PAM is directly used as the tendon connected to the fingers. On the other end of the PAM, we insert the air supply before closing the knot, consisting of a PTFE tube with an inner diameter of \qty{0.8}{\milli\meter} and an outer tube diameter of \qty{1.6}{\milli\meter}. The insertion length is \qty{5}{\milli\meter}. The air supply tube is connected to the pneumatic source.
The material cost for this PAM amounts to \qty{3.45}{USD/\meter}, creating a cost-efficient and easily replicable alternative to commercially available PAM.

\begin{figure}[t]
    \includegraphics[width=\columnwidth]{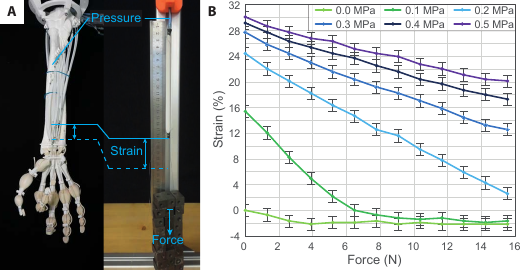}
    \centering
    \caption{\textbf{Experimental results for the SRL PAM.} \textbf{A}~The SRL PAMs used in the printed hand and forearm are tested in a force strain test setup. A weight is applied to simulate forces. Displacements can be recorded for different pressures applied to the PAM. \textbf{B}~Force strain curves for different pressures applied to the PAMs. The error indicators represent an approximated measurement inaccuracy corresponding to \qty{2}{\milli\meter} stroke.}
    \label{Fig:SRL-PAM}
\end{figure}

We performed load-lifting experiments with incrementally increased pressure to characterize the force-strain behavior of our PAM (\Cref{Fig:SRL-PAM}). The force increases approximately linearly with increasing operating pressures. The maximum force measured at \qty{0}{\percent} strain was \qty{38.05}{\newton} at \qty{0.5}{\mega\pascal}. The maximum strain measured was \qty{30.1}{\percent} at \qty{0.5}{\mega\pascal}, which was just slightly lower than the highest McKibben strain reported in literature~\cite{kurumaya_design_2017}.

\subsection{Pneumatic touch sensors}

\begin{figure}[t]
    \includegraphics[width=\columnwidth]{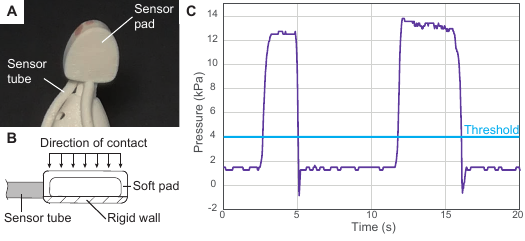}
    \centering
    \caption{\textbf{Pneumatic touch sensor.} \textbf{A}~Schematic of a pressure sensor pad with tube connection. \textbf{B}~Photo of the printed functional pressure sensor pad with tubing. \textbf{C}~Sensor pressure over time showing several contacts with the sensor pad.}
    \label{Fig:PressureSensor}
\end{figure}

To provide the hand with a sense of touch, we integrated sensor cavities in the fingertips and the palm (\Cref{Fig:PressureSensor}). These cavities were printed from a low-viscosity polymer and are used to relay a fingertips’ and palm's contact with an object. When touching an object with sufficient force, the pouch is compressed, leading to a measurable pressure increase. The sensors and connected tubes are integrated into the hand component and fully printed. 
%A hole in the sensor back wall is required to remove the wax support material. This hole is sealed with hot glue to make the sensor airtight. 
A printed soft tube connects the sensor cavity with a pressure sensor\footnote{Honeywell SSCMANV015PG2A3}. 
We used the pressure readings for valve control by comparing them to a pre-defined threshold of \qty{4}{\kilo\pascal}.

This threshold can be adjusted depending on the desired tightness of the grip. When the pressure of the touch sensor of one finger exceeds the threshold during a flexion motion, the lead pressure increase for the flexor PAMs of this finger is stopped, and the lead pressure is held constant at the last value. This control strategy aims at adaptive grasping of objects, in which each finger bends until a touch contact is measured so the finger positions adjust individually to an unknown object to grasp it.

\subsection{Fingertip and Grasping Force Test Setup}
\label{secc:force_test}
\begin{figure}[t]
    \includegraphics[width=\columnwidth]{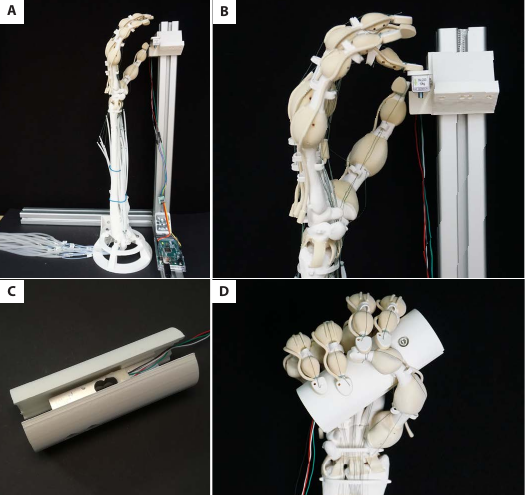}
    \centering
    \caption{\textbf{Setup for force measurements.} \textbf{A}~Fingertip force test, load cell connected to a microcontroller for readout. \textbf{B}~A fingertip applying force to the load cell. \textbf{C}~Load cell in between two cylinder halves used to measure the grasping force compressing the cylinder. \textbf{D}~Grasping force test of the robotic hand with the test object shown in sub-figure C.}
    \label{Fig:GraspingForce}
\end{figure}

We characterized the force the hand can exert by 
measuring the force an individual finger can apply with a load cell (\Cref{Fig:GraspingForce}A, B). We also tested the hand's grasping force by constructing a cylinder of two half-cylinders with a load cell connecting the two halves (\Cref{Fig:GraspingForce}C, D). The load cell measures the force of grasping and thereby compressing the cylinder during a medium wrap power grasp~\cite{feix_grasp_2016}.

\section{Results}

\subsection{Range of Motion}
\begin{figure}[ht]
    \includegraphics[width=\columnwidth]{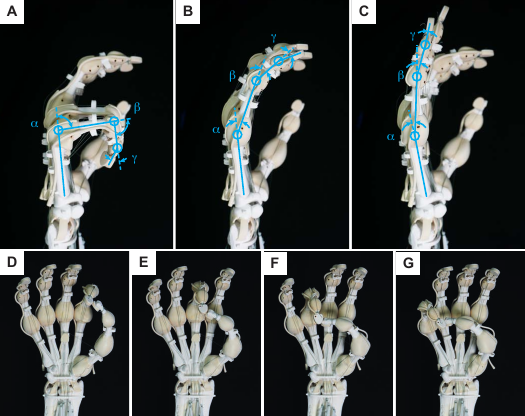}
    \centering
    \caption{\textbf{Range of motion analysis of the printed hand.} The images \textbf{A}, \textbf{B}, and \textbf{C} show the fingers of the hand in \textbf{A} flexed, \textbf{B} relaxed, and \textbf{C} extended position. The joint angles were visually approximated from still images. A Kapandji test, touching the fingertip of each finger with the thumb, is shown in sub-figures \textbf{D} through \textbf{G}.}
    \label{Fig:ROM}
\end{figure}

The range of motion of the finger joints is analyzed for the little finger (\Cref{Fig:ROM}A-C and \Cref{tab:ROM}) by measuring the finger joint angles in a stretched, relaxed, and curled position. The flexor PAMs are operated at a lead pressure of \qty{0.2}{\mega\pascal}. The denoted range is not the physical maximum, but further tendon displacement could lead to damage to the prototype.

\begin{table}[ht]
    \centering
    \caption{Range of Motion.}
    \begin{tabular}{cccc}\toprule
      & \textbf{Stretched} & \textbf{Relaxed} & \textbf{Curled} \\\midrule
    $\alpha$ & \ang{7.2}      & \ang{29.7}    & \ang{90.9}   \\
    $\beta$ & \ang{15.0}      & \ang{29.8}    & \ang{87.1}   \\
    $\gamma$ & \ang{0.0}       & \ang{14.5}    & \ang{45.0}  
    \\\bottomrule
    \end{tabular}
    \label{tab:ROM}
\end{table}

\subsection{Dexterity and grasping}
We performed a Kapandji test~\cite{kapandji_cotation_1986} to assess the opposability of the thumb. In this medical test, the thumb touches the fingertips or joints of the long fingers and is attributed a score from 1 to 10. The system successfully touched all fingertips of the long fingers with the thumb, corresponding to a Kapandji score of 6 (see \Cref{Fig:ROM}D-G). The thumb could not reach the joints of the little finger. While this score is lower than the ideal human performance, the experiment demonstrates a good opposability of the thumb, which is on the same level as highly anthropomorphic robotic hands~\cite{grebenstein_hand_2012}.

\begin{figure}[t]
    \includegraphics[width=\columnwidth]{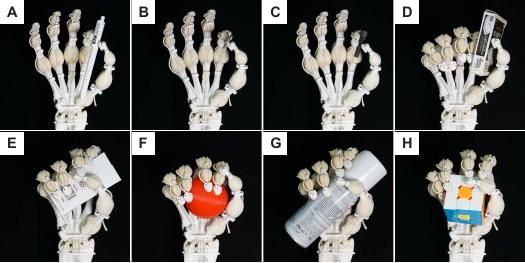}
    \centering
    \caption{\textbf{Grasping different objects.} We demonstrate the ability to grasp a variety of objects firmly. Pinch grasps for \textbf{A}~a ballpoint pen, \textbf{B}~a coin, \textbf{C}~a screw, and \textbf{D}~a pack of cards. An enveloping grasp is shown for \textbf{E}~a cardboard box, \textbf{F}~a ball the size of a tennis ball, \textbf{G}~a spray can, and \textbf{H}~a Rubik's cube.}
    \label{Fig:Grasps}
\end{figure}

Various finger motions, hand gestures, and grasps were tested with this system (\Cref{Fig:Grasps}). We realized power grasps, precision grasps, and intermediate grasps~\cite{feix_grasp_2016} of a large variety of differently shaped objects with the hand. The object set included a pen, a coin, a screw, cardboard boxes, plastic bottles, a spray can, a sphere of \qty{67}{\milli\meter} diameter (corresponding to a tennis ball), a Rubik’s cube and other objects with weights of up to \qty{272}{\gram}. A full flexion and extension movement -- closing and opening the hand -- takes around \qty{3}{s}. Higher speeds are possible but have not been tested.

\subsection{Fingertip and Grasping Forces}
We measured the fingertip and grasping forces with the test setup described in \Cref{secc:force_test}. \Cref{tab:Forces} shows the results. The measured fingertip force was \qty[separate-uncertainty=true]{1.95(0.15)}{\newton} at a PAM operating pressure of \qty{0.2}{\mega\pascal}. 
We measured a grasping force of \qty[separate-uncertainty=true]{2.97(0.25)}{\newton} at a lead pressure of \qty{0.18}{\mega\pascal}. We estimate a lower systematic error for the fingertip force compared to the grasping force test. Because we used a 1D load cell in the test object, slight deviations in its orientation when grasping can effect the result.  For our human participant, we measured \qty{24}{\newton} at subjectively moderate effort and \qty{43}{\newton} at high effort.

\begin{table}[ht]
    \centering
    \caption{Fingertip and Grasping Forces.}
    \begin{tabular}{ccc}\toprule
         \textbf{Type} & \textbf{Force} & \textbf{Actuator Pressure} \\\midrule
         Fingertip & \qty[separate-uncertainty=true]{1.95(0.15)}{\newton} & \qty{0.2}{\mega\pascal} \\
         Grasping & \qty[separate-uncertainty=true]{2.97(0.25)}{\newton} & \qty{0.18}{\mega\pascal} \\
         Human moderate grasp & \qty{24}{\newton} & N/A \\
         Human strong grasp & \qty{43}{\newton} & N/A
                  \\\bottomrule
    \end{tabular}
    \label{tab:Forces}
\end{table}

\section{Discussion}

In this work, we developed and integrated a cutting-edge design and fabrication method of a highly biomimetic robotic hand and forearm system featuring state-of-the-art soft pneumatic artificial muscles based on the McKibben PAMs. Simplifying the fabrication process by integrating multi-material 3D printing pushes the boundaries of biomimetic hand design. It allows us to quickly and accurately manufacture and test entire designs made of different soft and rigid materials. Furthermore, touch sensors based on pressure readings can be directly printed and integrated into the design.

Our pneumatic artificial muscles achieved up to \qty{30.1}{\%} strain and forces as high as \qty{38.05}{\newton}, matching state-of-the-art McKibben muscles at a fraction of their material cost while decreasing the manufacturing effort. The 22 muscles actuate the tendon-driven hand, fabricated as a single-part multi-material 3D print. This innovative design facilitates flexion and extensions in all fingers and enables a range of dexterous hand gestures and movements. 

We showcased a high range of motion and the possibility of grasping very different objects. The integration of an opposable thumb proved to be crucial. In the Kapandji test, the hand reached a score of 6 out of 10. While this score is high for robotic hands, it lacks behind the human inspiration. The measured maximal fingertip force was \qty{1.95}{\newton}, with material failure being the primary limiting factor.
This complete system weighs \qty{431}{\gram} and not only mirrors the appearance of a human hand but also ensures ease of assembly.

The material strength of our printed components is significantly lower than their biological counterparts. For instance, our printed rigid material has an ultimate tensile strength of 45 MPa, starkly contrasting with the \qty{133}{\mega\pascal} exhibited by the cortical bone. Furthermore, we observed wear effects on joint capsules and articulating bone surfaces after approximately 50 flexion movements of the hand, indicating potential durability issues. The pneumatic pumps and valves required for the PAMs also presented challenges, being bulky and expensive, limiting the system’s mobility.

\section{Conclusion}

We presented a highly dexterous and functional robotic hand and forearm system based on human MRI scans. Vision controlled jetting allowed us to drastically speed up the fabrication process and fully 3D print a soft-rigid robot. Our easy-to-manufacture, cost-effective, and high-strain pneumatic artificial muscle provides muscle-like actuation.

Future work will focus on refining the design for enhanced grasping, improving the touch sensors, and addressing the issues of joint durability and finger kinematics. Integrating additional sensors and implementing advanced feedback control systems will be crucial in achieving more precise control and functionality. Additionally, exploring untethered compressed air sources could significantly enhance the system’s mobility and application range, bringing us closer to realizing a fully functional robotic arm.

\addtolength{\textheight}{-4cm}   % This command serves to balance the column lengths

%%%%%%%%%%%%%%%%%%%%%%%%%%%%%%%%%%%%%%%%%%%%%%%%%%%%%%%%%%%%%%%%%%%%%%%%%%%%%%%%

%%%%%%%%%%%%%%%%%%%%%%%%%%%%%%%%%%%%%%%%%%%%%%%%%%%%%%%%%%%%%%%%%%%%%%%%%%%%%%%%
\section*{ACKNOWLEDGMENT}
The authors thank the team at Inkbit Inc. for printing the hand structure, for their support, and for their advice when designing and operating the robotic structures. We are also grateful for the valuable advice from Stefan Weirich's advisors at RWTH Aachen and the support by the IDEA League scholarship for Stefan Weirich.

%%%%%%%%%%%%%%%%%%%%%%%%%%%%%%%%%%%%%%%%%%%%%%%%%%%%%%%%%%%%%%%%%%%%%%%%%%%%%%%%

\printbibliography
% \bibliographystyle{IEEEtran}
% \bibliography{IEEEabrv,RoboSoft24VCJ-AK}

\end{document}